\documentclass[conference]{IEEEtran}
\usepackage{amsmath,amssymb,amsfonts}
\usepackage{algorithmic}
\usepackage{graphicx}
\usepackage{textcomp}
\usepackage{xcolor}
\usepackage{balance}
\usepackage{tabularx}  
\usepackage{booktabs}  
\usepackage{fancyhdr}
\usepackage{url}
\usepackage[square,comma,sort&compress,numbers]{natbib} 
\usepackage{threeparttable,multirow,adjustbox,threeparttable}
\usepackage[symbol]{footmisc}
\usepackage{orcidlink} 

\usepackage{pbalance}  
\usepackage{hyperref}
\hypersetup{
    colorlinks=true,        
    linkcolor=blue,         
    citecolor=blue,         
    filecolor=blue,         
    urlcolor=blue           
}
\usepackage{caption}
\begin{document}
\title{Multi-Slice Spatial Transcriptomics Data Integration Analysis with STG3Net}
\author{
\IEEEauthorblockN{Donghai Fang$^1$, Fangfang Zhu$^2$ and Wenwen Min$^{1*}\orcidlink{0000-0002-2558-2911}$  }\\
\IEEEauthorblockA{
$^1$School of Information Science and Engineering, Yunnan University, Kunming 650091, Yunnan, China \\
$^2$College of Nursing Health Sciences, Yunnan Open University,  Kunming 650599, Yunnan, China\\
$^{*}$Correspondence author. 
Contact: \href{email}{fangdonghai@aliyun.com}, \href{email}{minwenwen@ynu.edu.cn}
}				
}
\maketitle
\thispagestyle{fancy}
\begin{abstract}
With the rapid development of the latest Spatially Resolved Transcriptomics (SRT) technology, which allows for the mapping of gene expression within tissue sections, the integrative analysis of multiple SRT data has become increasingly important. However, batch effects between multiple slices pose significant challenges in analyzing SRT data. To address these challenges, we have developed a plug-and-play batch correction method called Global Nearest Neighbor (G2N) anchor pairs selection. G2N effectively mitigates batch effects by selecting representative anchor pairs across slices. Building upon G2N, we propose STG3Net, which cleverly combines masked graph convolutional autoencoders as backbone modules. These autoencoders, integrated with generative adversarial learning, enable STG3Net to achieve robust multi-slice spatial domain identification and batch correction. We comprehensively evaluate the feasibility of STG3Net on three multiple SRT datasets from different platforms, considering accuracy, consistency, and the F1LISI metric (a measure of batch effect correction efficiency). Compared to existing methods, STG3Net achieves the best overall performance while preserving the biological variability and connectivity between slices. Source code and all public datasets used in this paper are available at Github (\url{https://github.com/wenwenmin/STG3Net}) and Zenodo (\url{https://zenodo.org/records/12737170}).
\end{abstract}


\section{Introduction}
In complex organisms, cells are organized into specialized clusters with distinct functions through dynamic interactions and intricate arrangements among them \cite{evaluation}. These clusters exhibit mutual influence and close connections to collectively carry out the coordinated operation of the organism. The latest spatially resolved transcriptomics (SRT) techniques \cite{10xvisium, stereo}, such as ST with a spatial resolution of 100 $\mu m$, 10x Visium with a spatial resolution of 55 $\mu m$, and Stereo-seq with a spatial resolution of 220 $nm$, enable genome-wide analysis at the multicellular or even single-cell level, capturing the transcriptional expression corresponding to specific spatial locations (referred to as spots) \cite{DLPFC, AMB, ME}. The insights into genomic transcriptional expression based on spatial information provide a solid foundation for researchers to understand many biological processes that affect diseases and facilitate advancements in the diagnosis and treatment of related conditions \cite{min2024dimensionality, STAGATE}.

An important computational task in recent SRT data analysis is the identification of shared and specific clusters, referred to as spatial domains, which are defined as regions with similar spatial expression patterns. Several methods have been proposed to address the spatial domain identification problem in single-slice SRT datasets \cite{SpaGCN,DeepST,STAGATE,min2024dimensionality,BIBM1,BIBM2} and data denoising \cite{li2024stmcdi, spadit}. For example, SpaGCN \cite{SpaGCN} utilizes unsupervised clustering algorithms to detect different spatial expression patterns in single-slice SRT datasets. DeepST \cite{DeepST} further extracts information from histological images using convolutional networks and then employs a deep generative network to identify spatial domains. STAGATE \cite{STAGATE} utilizes a deep autoencoder model and employs graph attention mechanism to learn spatial heterogeneity. STMask \cite{min2024dimensionality} employs a dual-channel autoencoder with masks to learn the latent representation of a single slice. However, these methods face challenges in the comprehensive analysis of multiple SRT datasets since batch effects among multi-slices generated under different conditions, techniques, or developmental stages may obscure actual biological signals \cite{STAligner, Harmony}. This limitation poses a significant challenge to existing methods developed based on single-slice approaches and restricts our ability to understand the biological processes occurring between slices.

To address these challenges, several methods and tools have been developed for the integrated analysis of multiple SRT datasets. PASTE \cite{PASTE} assumes the maximum possible overlap between multiple slices and improves clustering performance by aligning neighboring slices to the central slice using alignment algorithms. However, its applicability is limited to cases in which slices only partially overlap or do not overlap spatially at all. Splane \cite{Splane} utilizes the Spoint model for deconvolution analysis of SRT data. The obtained cell type composition replaces the original transcriptional expression as the input to the model. By integrating adversarial learning strategies, it demonstrates the elimination of batch effects across slices. However, it is a multi-stage model, and the performance of spatial domain identification across slices is highly dependent on the deconvolution performance. STAligner \cite{STAligner} combines STAGATE with a method based on mutual nearest neighbors (MNNs) \cite{MNN} into a unified model, enabling the spatial awareness of multiple SRT datasets. However, the MNN-based method forces the search for MNN-pairs between arbitrary slices. In reality, slices may not necessarily contain the same specialized clusters, such as at different time points, leading to the incorrect construction of MNN-pairs in different spatial domains. Additionally, computing MNN-pairs can be computationally intensive, potentially overlooking non-MNN-pairs within the same functional clusters. STitch3D \cite{STitch3D} aligns two-dimensional slices using the Iterative Closest Point (ICP) or PASTE algorithm and constructs a global three-dimensional spatial adjacency graph structure. It incorporates slice- and spot-specific effects as well as slice- and gene-specific effects into a graph attention autoencoder model to mitigate batch effects across slices. However, it also relies on cross-slice coordinate alignment, limiting its application.

In this study, we propose STG3Net, a deep learning framework for multi-slice spatial domain identification and batch correction. It utilizes a masked graph autoencoder as a self-supervised backbone, learning latent representations by reconstructing the expressions of masked spots to mitigate the occurrence of discrete spots. By employing generative adversarial learning, the model’s spatial domain identification capabilities are effectively enhanced, particularly in complex, multiple SRT datasets. Additionally, we develop a plug-and-play batch correction method called Global Nearest Neighbor (G2N), specifically designed for SRT datasets. Leveraging global semantic information, G2N captures a more comprehensive set of anchor nodes within the latent space.

The main contributions of our proposed method are:
\begin{itemize}
    \item We introduce a plug-and-play batch correction method, G2N, specifically designed for multiple SRT datasets and effectively mitigates batch effects.

    \item We cleverly integrate a masked graph autoencoder with adversarial learning to achieve a more robust multi-slice spatial domain identification. 

    \item STG3Net achieved the best overall performance on three different platform datasets and maintained homogeneity and variability among multi-slices.
\end{itemize}

\begin{figure*}[t]%
\centering
\centering
\includegraphics[width=1\textwidth]{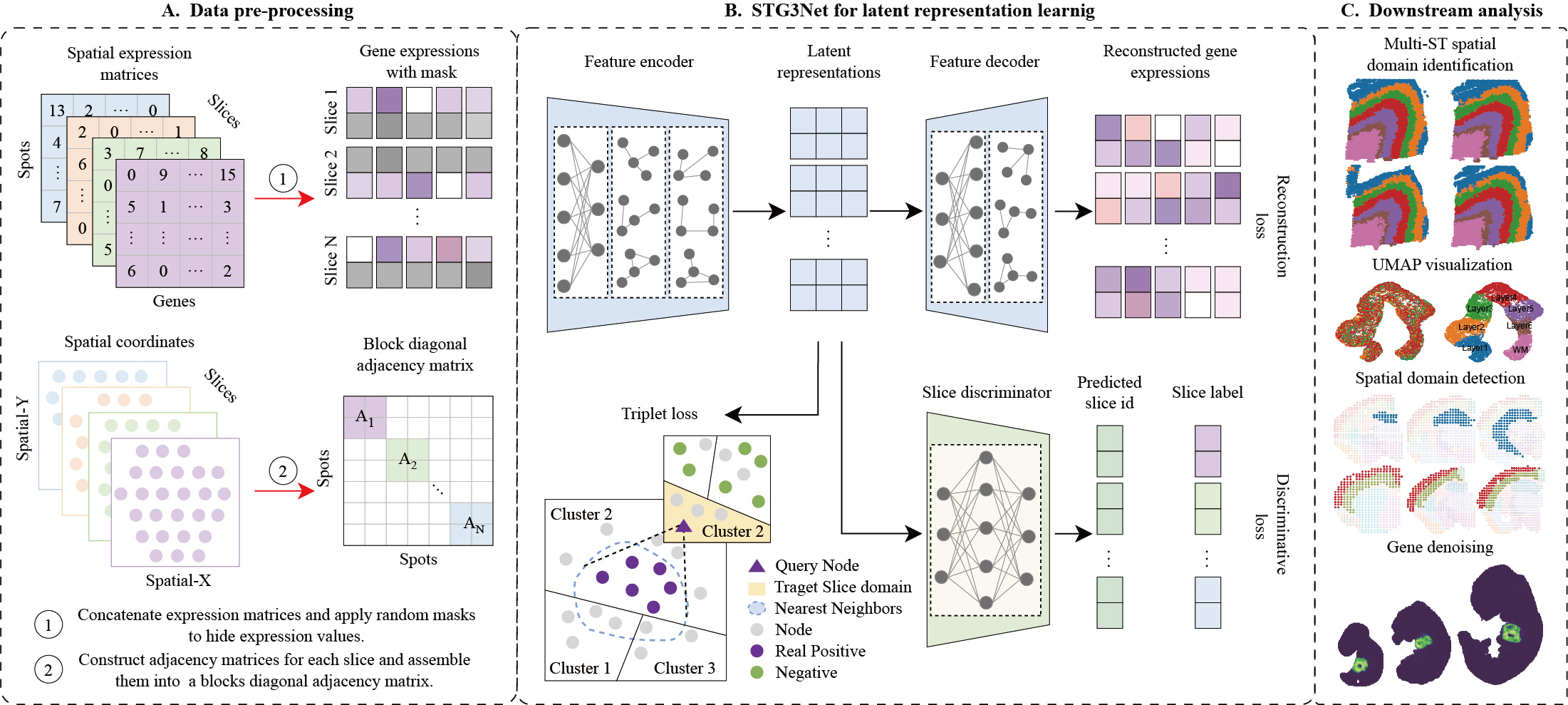}
\caption{Overview of STG3Net. (A) Data preprocessing involves integrating multiple SRT data, enhancing spot features, and constructing spatial adjacency graphs. (B) STG3Net is employed for latent representation learning. It consists of a backbone module composed of feature graph autoencoder, combined with adversarial learning and G2N for batch correction between multiple slices. (C) The learned latent representations from STG3Net will be utilized in downstream task analysis, including clustering and UMAP visualization. Additionally, the reconstructed gene expression is considered as the denoised outcome.}
\label{fig1}
\end{figure*}

\section{PROPOSED METHODS}\label{MATERIALS AND MRTHODS}
\subsection{Overview of the proposed STG3Net}
To start, the gene expression matrices from multiple slices are merged into a unified input matrix. Subsequently, masking operations refine specific features, thereby augmenting the data, while constructing a spatial neighborhood graph based on spatial positions. The acquired latent representation of gene expression from STG3Net is then leveraged for downstream tasks like multi-slice spatial identification. STG3Net integrates a graph autoencoder with masks, a slice discriminator, and G2N anchor pairs selection to enhance the model's performance in capturing spatial relationships effectively (\autoref{fig1}).

\subsection{Data augmentation and construction of neighbor graph}
STG3Net takes gene expression and spatial coordinate information of multiple SRT slices as input. The gene expression matrices of each slice are concatenated along the spot dimension to construct a gene expression matrix for multiple slices. Then, using functions provided by the SCANPY library, the interested genes are retained and subjected to regularization. Finally, the top $N_t$ highly variable genes are selected as the preprocessed model input, denoted as $X\in \mathrm{R}^{N\times N_t}$, where $N$ represents the total number of spots across all slices. To overcome the ``identity transformation" issue, we use the augmented representation $\widetilde{X}\in \mathrm{R}^{N\times N_t}$ as the input for training the model. Specifically, from the set of spot vertices, a masked vertex set $\mathcal{V}_m$ is randomly sampled with a masking rate $\rho $. For the $i$-th spot ($v_i$), if $v_i \in \mathcal{V}_m$, $\widetilde{x}_i = x_{[mask]}$, where $x_{[mask]}$ represents the mask token vector; otherwise, $\widetilde{x}_i = x_i$.

For any given slice (indexed as $s$), the Euclidean distance between spots is calculated based on their spatial coordinates. The $K$-nearest neighbor (KNN) algorithm is then applied to select the top $K$ nearest neighboring spots, resulting in the construction of an adjacency matrix $A_s$. If $j$-spot is a neighbor of $i$-spot, then $A_{s,ij} = A_{s,ji} = 1$. Subsequently, the adjacency matrices of each slice are combined to form a block diagonal adjacency matrix $A$, which serves as the input.

\subsection{Latent representation learning with masked reconstruction}
The latent representation of gene expression is obtained from the encoder. The encoder first takes the masked matrix $\widetilde{X}$ and passes it through two stacked fully connected layers to generate a low-dimensional representation $H_f\in \mathrm{R}^{N\times d_f}$. 
Additionally, leveraging the advantages of the SRT dataset, a graph convolutional network (GCN) is utilized to process and aggregate information from neighboring nodes, facilitating the integration of gene expression and spatial coordinates. This results in a meaningful and useful latent representation $H\in \mathrm{R}^{N\times d}$. Here, $d_f$ and $d$ are the dimensions of the low-dimensional expression representation, and the final latent representation of STG3Net, respectively. The calculation of $H$ involves a two-layer GCN, which is defined as follows:
\begin{equation}
    H=\text{GCN}_{h}(A, H_f)=\widetilde{A}\text{ReLU}(\widetilde{A}H_fW_0)W_1
\end{equation}
with weight matrices $W_0$, $W_1$, and symmetrically 
normalized adjacency matrix $\widetilde{A}=D^{-\frac{1}{2}}AD^{-\frac{1}{2}}$. 

Once the training phase is completed, we utilize $X$ as the input to the encoder and obtain the latent representation $H$. This representation is then used for downstream tasks such as spatial domain identification and visualization in multiple SRT dataset analysis. 

The decoder consists of one fully connected layer and one GCN, which is used to reconstruct the gene expression $Z\in \mathrm{R}^{N\times N_t}$ from the latent representation. The calculation of $Z$ is as follows:
\begin{equation}
    Z=\text{GCN}_{z}(A, H)=\widetilde{A}\text{LeakyReLU}(H{W}'_0){W}'_1
\end{equation}
with weight matrices ${W}'_0$ and ${W}'_1$.

Therefore, one of the main objectives of STG3Net is to reconstruct the masked gene expression of spots in $\mathcal{V}_m$ given a partially observed spot set and adjacency relationships. By utilizing the Scaled Cosine Loss (SCE) as the objective function, it is defined under a predetermined scaling factor $\gamma$ as follows:
\begin{equation}
    \mathcal{L}_{sce}=\frac{1}{|\mathcal{V}_m|}\sum_{v_i\in\mathcal{V}_m}{(1-\frac{x_iz_i^{\mathsf{T}}}{\left\|x_i\right\|\left\|z_i\right\|})^{\gamma}} ,\gamma\ge 1
\end{equation}
where $\gamma$ is fixed to 2 to reduce the contribution from simple samples during the training process, and $|\mathcal{V}_m|$ represents the number of spots in the masked set. 

\subsection{Adversarial learning for multiple slices}
To improve the clustering performance across multiple slices, we have designed a discriminator consisting of three stacked fully connected layers. It takes the latent representation outputted by the encoder as input and produces the probability $p_{i,s}$ of $i$-spot belonging to $s$-slice. Therefore, under the supervision of the slice label $y_{i,s}$ of $i$-spot, we have the discriminator loss as follows:
\begin{equation}
    \mathcal{L}_{dis}=\frac{1}{|\mathcal{V}_m|}\sum_{v_i\in \mathcal{V}_m}{y_{i,s}\log (p_{i,s})}
\end{equation}

The $\mathcal{L}_{dis}$ is minimized when there is a need to discover distinguishing features between slices from the latent representation, allowing the discriminator to accurately predict the slice labels. Conversely, $\mathcal{L}_{dis}$ is maximized when there is a need to deceive the discriminator to mitigate batch effects across multiple slices, ensuring that spots from different slices would have the highest similarity. 

\subsection{Triplet learning with global nearest neighbors}
We have developed a novel Global Nearest Neighbor (G2N) anchor pairs selection method specifically for analyzing the multiple SRT dataset. This method utilizes the obtained G2N-pairs to overcome batch effects across multiple slices. Firstly, we define the anchor $\mathcal{A}_{s,i}$ as the $i$-th spot on the current $s$-th slice, while the positive anchor is defined as the spot on all other $t$-th slices except the $s$-th slice. Based on this, we further calculate the similarity between the latent representations of the anchor and positive anchors and retain the top $K_g$ nearest positive anchors for each anchor, denoted as $\mathcal{B}_{s,i}$. To further obtain a global set of positive anchors, we employ clustering to discover and acquire nodes that share global semantic information with the anchors. By utilizing a clustering algorithm, we cluster the spots into $K_c$ clusters and select nodes belonging to the same domain as the anchor as similar points that share semantic information globally. This set is denoted as $\mathcal{C}_{s,i}$. Therefore, we obtain the set $\mathcal{R}_{s,i}=\mathcal{B}_{s,i}\cap\mathcal{C}_{s,i}$ of globally nearest anchor nodes in the latent space. To prevent excessive reliance on cluster distribution and avoid mode collapse, we randomly select $n/2$ nodes from the set $\mathcal{R}_{s,i}$ as the final positive anchor set $\mathcal{P}_{s,i}$. The negative anchors are defined on the $s$-th slice, selecting $|\mathcal{P}_{s,i}|$ nodes from different domains than the current anchor as the negative anchor set $\mathcal{N}_{s,i}$. 

The triplet loss is employed to minimize the distance between anchor-positive pairs and maximize the distance between anchor-negative pairs in the latent space. The calculation is as follows:
\begin{equation}
\footnotesize{
    \mathcal{L}_{tri}=\frac{1}{N_{tri}}\textstyle\sum_{(a,p,n)\in\mathcal{T}}^{N_{tri}}\max(\left \| h_a-h_p \right \|_2 -\left \| h_a-h_n  \right \|_2+ \tau,0)  }
\end{equation}
where $\mathcal{T}$ is the set of all anchor points $\mathcal{A}_{s,i}$ along with their corresponding positive anchor set $\mathcal{P}_{s,i}$ and negative anchor set $\mathcal{N}_{s,i}$, forming the G2N-pairs. $N_{tri}$ represents the total number of G2N-pairs and $\tau$ is the margin (default 1.0).

Finally, the entire learning objective is written as:
\begin{equation}
    \mathcal{L}=\mathcal{L}_{sce}-\lambda \mathcal{L}_{dis} + (1-\lambda)\mathcal{L}_{tri}
\end{equation}
where $\lambda$ is a hyperparameter that control the contribution in the loss function.

\subsection{Evaluation criteria}
We utilize accuracy metrics to describe the clustering precision of the method, consistency metrics to measure the dispersion of clustering results \cite{evaluation}, and the local inverse Simpson index to evaluate the batch correction effect across multiple slices \cite{F1LISI}. Specifically, in accuracy description, we have the following metrics: Adjusted Rand Index (ARI), used to compare the similarity between clustering results and manually annotated labels. Normalized Mutual Information (NMI), based on information theory, it measures the normalized mutual information between clustering results and true labels. Homogeneity (HOM) score, a metric that assesses if all clusters contain only data points belonging to a single class, indicating homogeneous clustering. Completeness (COM) score, a metric that measures if all data points belonging to a particular class are grouped together in the same cluster, indicating complete clustering \cite{evaluation}. Therefore, the overall accuracy score is calculated as follows:
\begin{equation}
    \text{Accuracy} = \frac{1}{3} \times (\text{NMI} + \text{HOM} + \text{COM})
\end{equation}
The closer the ARI and Accuracy scores are to 1, the better the clustering precision. 

We assess spatial continuity using the Spatial Chaos Score (CHAOS), where a lower chaos value indicates better identified spatial domain continuity. The Percentage of Anomalous Points (PAS) represents the randomness of points located outside the clustered spatial domains, with a lower PAS score indicating better detected spatial domain continuity \cite{evaluation}. Therefore, a lower overall score for consistency corresponds to better continuity. The calculation is as follows:
\begin{equation}
    \text{Consistency} = \frac{1}{2} \times (\text{CHAOS} + \text{PAS})
\end{equation}

We evaluate the degree of separation between the same domain and different domains by using the Local Inverse Simpson Index (LISI) for each batch within a domain ($\text{LISI}\_\text{batch}$) and the LISI across all data for the domain ($\text{LISI}\_\text{domain}$). The F1 score of the LISI \cite{F1LISI} is calculated as follows:
\begin{equation}
    \text{F1}_\text{LISI} = \frac{2(1-\text{LISI}\_\text{domain}_{norm})(\text{LISI}\_\text{batch}_{norm})}{1-\text{LISI}\_\text{domain}_{norm}+\text{LISI}\_\text{batch}_{norm}}
\end{equation}
A higher F1 score indicates superior batch correction.

\begin{figure*}[!t]%
\centering
\includegraphics[width=1\textwidth]{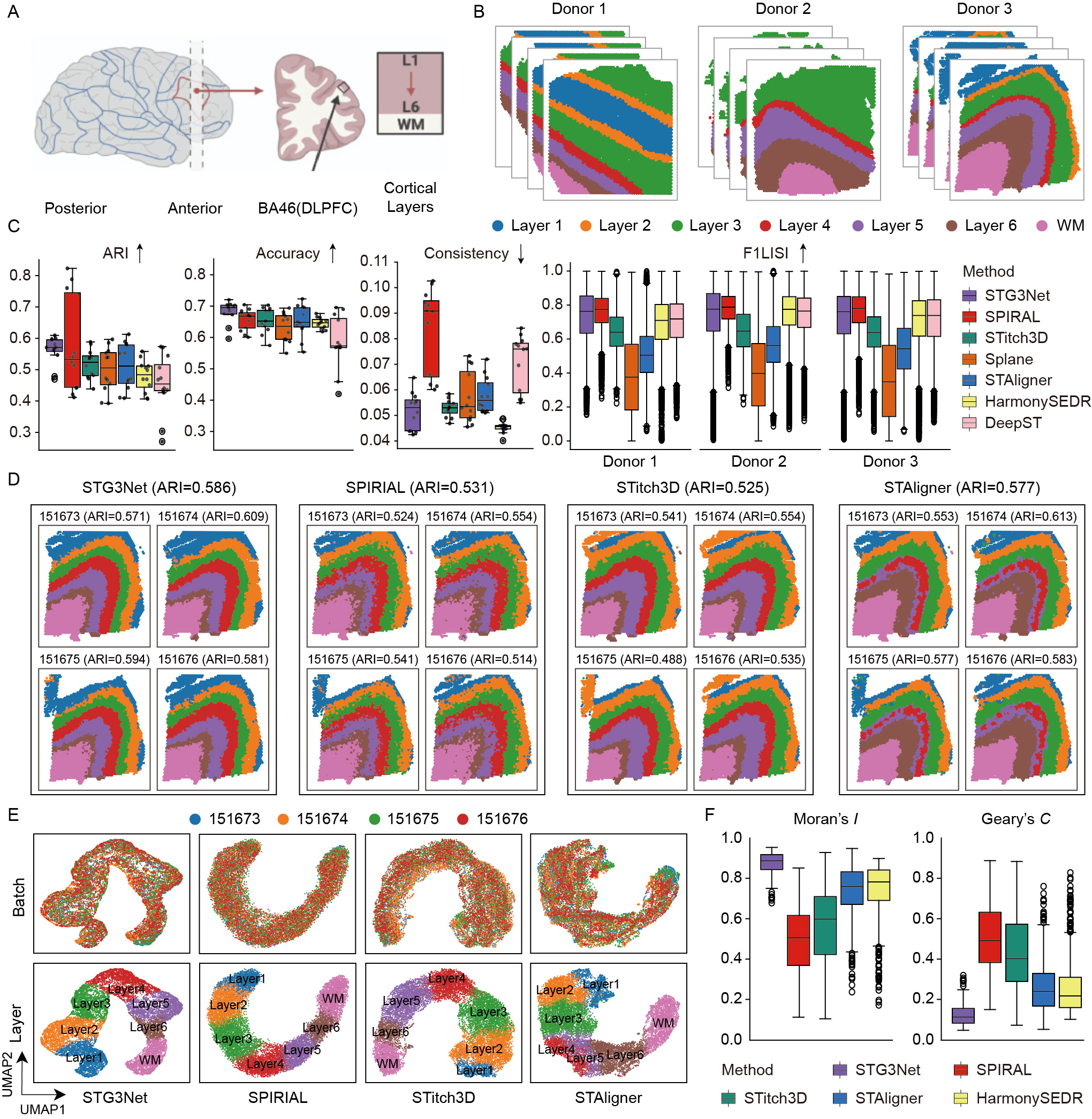}
\caption{DLPFC. (A) Schematic of DLPFC data. (B) Manual annotations for Donor 1, 2, and 3. (C) The evaluation of STG3Net and existing methods on the DLPFC dataset was conducted based on consistency, accuracy, and the F1LISI metric. (D) We conduct a comparison of spatial domain identification on four slices from the Donor 3 sample. (E) The UMAP plot of slice embeddings colored by batch and cortical layers. (F) Imputed gene spatial autocorrelation.}
\label{fig2}
\end{figure*}

\section{EXPERIMENTS}\label{result}
\subsection{Dataset description}
In this study, we analyze multiple SRT datasets from three different platforms (\autoref{tab-1}). Firstly, we examine the human dorsolateral prefrontal cortex (DLPFC) \cite{DLPFC} dataset obtained from the 10x Visium platform. This dataset consists of three independent donors, with each donor having four slices. Each slice is manually annotated with 5 to 7 spatial domains. Secondly, we consider an adult mouse brain (AMB) \cite{AMB} dataset from the ST platform, which includes 35 slices and up to 15 annotated spatial domains. Lastly, we investigate mouse embryo (ME) \cite{ME} slices at different developmental stages from the Stereo-seq platform. We select slices from the E9.5, E10.5, and E11.5 time periods. 

\begin{table}[t]
\caption{The statistics of the datasets used in this study.}\label{tab-1}
\begin{adjustbox}{width=0.5\textwidth}
\begin{tabular}{l|c|c|c|c|c}
\hline
Datasets      & \#Slices  & \#Spots  & \#Genes   &  \#Domains  & Platform                  \\ \hline
DLPFC \cite{DLPFC}   & 12   & 3,460-4,789 & 33,538    & 5-7  & 10x Visium    \\
AMB \cite{AMB}        & 35  & 152-620 & 23,371     & 3-12  & ST       \\
ME   \cite{ME}      & 3  & 5,913-30,124 & 22,385-26,854     & 12-19   & Stereo-seq     \\ \hline
\end{tabular}
\end{adjustbox}
\end{table}

\begin{figure*}[!t]%
\centering
\includegraphics[width=1\textwidth]{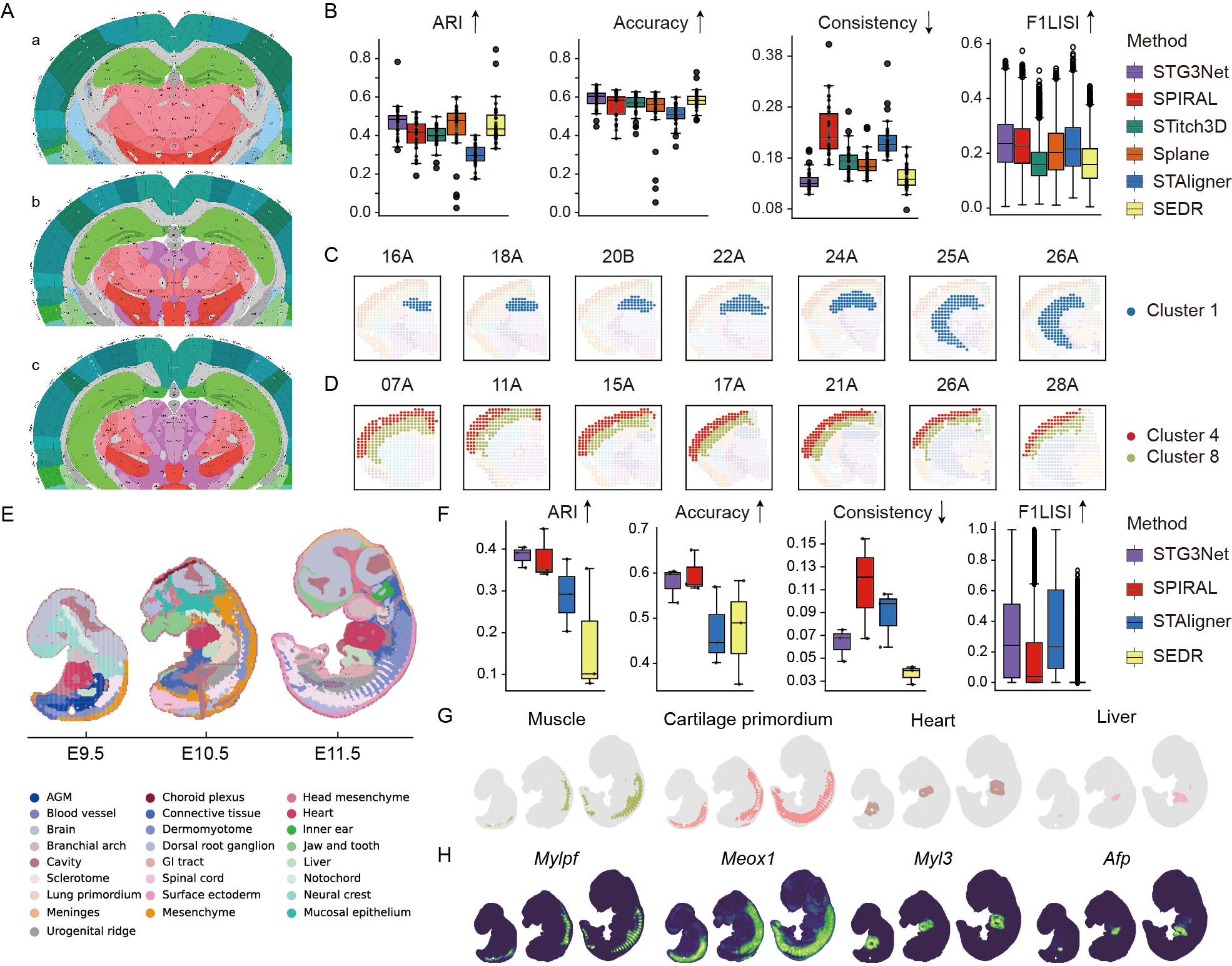}
\caption{AMB and ME. (A) Mouse brain coronal anatomical reference. a-c. (B) The evaluation of STG3Net and existing methods on the AMB dataset was conducted based on consistency, accuracy, and the F1LISI. (C) Cluster 1 corresponds to the Hippocampal region and exhibits features that preserve the developmental variability between slices along the AP axis. (D) The connectivity between Isocortical regions. (E) Manual annotation of tissue regions during the developmental stages of ME, including E9.5, E10.5, and E11.5. (F) The evaluation of STG3Net and existing methods on the ME dataset. (G) and (H) Denoised marker gene expression profiles for the developmental processes of four major tissues and organs, along with the corresponding annotated regions. }
\label{fig-3}
\end{figure*}

\subsection{Baseline methods}
We selected several state-of-the-art methods that are representative of the following:
\begin{itemize}
    \item SPIRAL \cite{SPIRAL} utilizes the GraphSAGE model to learn the latent representation of SRT data. This model divides the representation into two parts and employs a discriminator and a classifier to correct for batch effects. 
    \item STitch3D \cite{STitch3D} utilizes ICP or PASTE algorithm to maximize the overlap between multiple slices. It models the slice-spot factor and slice-gene factor, and reconstructs gene transcription expression using cell composition components guided by scRNA-seq data. 
    \item Splane \cite{Splane} utilizes Spoint to obtain the composition components of marked cells from scRNA-seq data, which are then used as inputs for Splane. Splane leverages adversarial learning to address the batch issue in multiple slices. 
    \item STAligner \cite{STAligner} leverages STAGATE to learn the latent representation and identifies MNN-pairs from the latent space. It computes the triplet loss.
    \item SEDR \cite{SEDR} utilizes a deep autoencoder network to learn gene expression while simultaneously incorporating the spatial information from the variation graph autoencoder.
    \item DeepST \cite{DeepST} employs GCN as an encoder to reconstruct the input graph topology and capture the point features for spatial domain identification.
\end{itemize}

\subsection{Implementation Details}
For STG3Net, we use a learning rate of 0.001 and a weight decay of 2e-4, optimized with Adam. The encoder have linear layers of dimensions 64 and 32, followed by GCN layers with output dimensions of 64 and 16. The feature decoder reconstructs to the raw data space. For all baselines, default parameters from the raw papers were used, and all experiments were conducted on an NVIDIA GeForce RTX 3090.

\subsection{STG3Net achieves consistent integration among the donors of DLPFC dataset}
We applied STG3Net to the DLPFC dataset (\autoref{fig2}A), using each independent donor (\autoref{fig2}B) as an input unit and calculated F1LISI. We then computed the accuracy and consistency of each slice based on the results obtained from all donors \cite{DLPFC}. From Figure 2C, we observed that STG3Net achieved the best median ARI and Accuracy values across 12 slices, indicating its good clustering accuracy in multi-slice spatial domain identification. We noted that although SPIRAL had the best F1LISI values across the three donors, it exhibited the poorest consistency. Examining the identified spatial domains (Figure 2D), SPIRAL had a significant number of discrete spots, further proving its inability to achieve good consistency. SPIRAL divided the latent representation into two parts, one aiming to learn inter-slice differences and the other aiming for maximum inter-slice mixing. This allowed it to achieve impressive batch correction performance but also contributed to its poor consistency.

Both STG3Net and SEDR involved randomly masking a portion of spot expressions, forcing the model to learn meaningful representations from neighbors. This ensured that the model obtained good consistency because the masking technique allowed the model to focus more on the current spot's relationship with its neighbors. Splane did not perform well in batch correction, as demonstrated by our ablation study. We also found that simple adversarial learning could improve multi-slice clustering accuracy to some extent,  but with limited performance.

Both STG3Net and STAligner relied on triplet loss to overcome batch effects by reducing the distance between anchor and positive anchors. Although MNN has been proven effective in removing batch effects in single-cell datasets, it is not entirely suitable for SRT data. MNN enforced the identification of MNN-pairs between every pair of slices, but in actual SRT data, not every pair of slices corresponded to the same spatial domain. This would erroneously construct MNN-pairs across different spatial domains. Additionally, MNN required computing mutual nearest neighbors, which could overlook non-MNN-pairs from the same spatial domain. On the other hand, STG3Net used our developed G2N positive anchor selection method, overcoming the aforementioned issues. From the UMAP plot of slice embeddings colored by cortical layers (Figure 2E), it can be seen that STG3Net effectively alleviated batch effects between slices.

In each identified spatial domain, we detected the top 100 differentially expressed genes ($|\log 2\text{FoldChange}|\ge 2$ and $\text{P}_{value}<0.05$) as mark genes. Subsequently, we calculated the Moran's $I$ and Geary's $C$ statistics \cite{evaluation} for the denoised gene expression (Figure 2F). From the results, it can be seen that STG3Net exhibited high spatial autocorrelation and tended to cluster together. 

\subsection{STG3Net preserves the biological variability and connectivity across adult mouse brain (AMB) data}
We applied STG3Net to a complex adult mouse brain dataset (\autoref{fig-3}A) comprising 35 coronal slices spanning the anterior-posterior (AP) axis \cite{AMB}. The dataset exhibits varying sizes and categories of spatial domains along the AP axis, presenting challenges for conventional batch removal techniques. Our results showcased that, in comparison to alternative methods, STG3Net excelled in accuracy, consistency, and batch correction capabilities (\autoref{fig-3}B). Notably, the MNN method faced limitations when handling SRT datasets, particularly struggling with the complexity of datasets composed of numerous slices such as in this study. Its processing speed notably lagged behind, attributed to the computational intensity of computing mutual nearest neighbors for every slice pair, resulting in $C_n^2$ calculations for a dataset with $n$ slices. In contrast, G2N demonstrated efficiency by requiring only $n$ calculations.

The distribution of cluster 1 aligns accurately with the annotated region of the Hippocampus in the Allen Reference Atlas - Mouse Brain \cite{MouseBrain}. Throughout the developmental trajectory, STG3Net effectively preserved the biological variations present in complex slices (\autoref{fig-3}C). Notably, our analysis revealed that the Isocortex region, characterized by two laminar domains from cluster 4 and cluster 8, maintained consistent connectivity across the developmental trajectory, showcasing uniform cortical generation patterns across all slices (\autoref{fig-3}D) within the context of lung cancer research.

\subsection{STG3Net identifies the structural organization of developing mouse embryonic (ME) data}

We applied STG3Net to mouse embryo slices obtained from three distinct developmental stages: E9.5, E10.5, and E11.5 (\autoref{fig-3}E). Despite the presence of numerous distinct tissue domains and noticeable batch effects, STG3Net demonstrated remarkable clustering accuracy (\autoref{fig-3}F). Our analysis focused on the detection results of major tissues and organs, including Muscle, Cartilage primordium, Heart, and Liver, by STG3Net (\autoref{fig-3}G), revealing a high level of consistency with annotated regions \cite{ME_annotation}. Further examination unveiled that, throughout the developmental stages, the proportion of the heart remained relatively stable, while the liver displayed developmental changes over time. This observation underscores STG3Net's efficacy in integrating information from multiple slices to detect heterogeneity within tissue structures. Moreover, by selecting marker genes from these specific regions, we illustrated the expression patterns post-denoising (\autoref{fig-3}H). These genes, validated by existing literature, establish their association with relevant diseases or tissue development during mouse development, contributing valuable insights to lung cancer research.

\subsection{Ablation studies}
\begin{figure}[t]%
    \centering
    \includegraphics[width=0.5\textwidth]{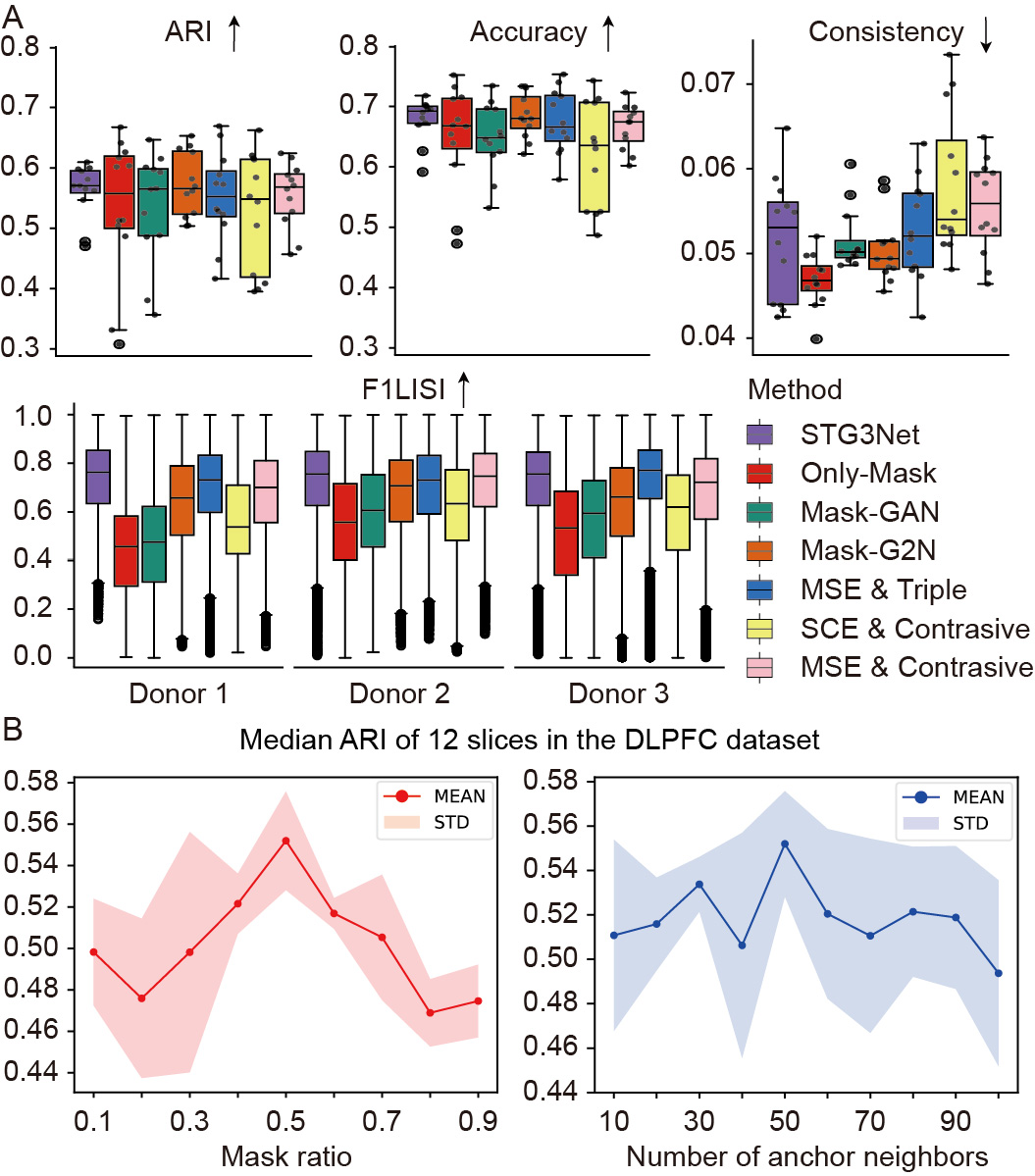}
    \caption{(A) Ablation studies of different components and objective functions. (B) Ablation studies of different mask rates and anchor node numbers.}\label{fig-4}
\end{figure}
We delved into the impact of various components on the model using three distinct variants. The first variant, Only-Mask, leveraged a feature autoencoder with a masking mechanism to enhance performance. The second variant, Mask-GAN, integrated adversarial learning for multi-slice spatial identification. The third variant, Mask-G2N, combined G2N batch correction with the masking mechanism. Results obtained from the DLPFC dataset (\autoref{fig-4}A) revealed that the Only-Mask variant excelled in continuity and accuracy, emphasizing neighboring information to suppress the discrete distribution of spots. However, it lacked multi-slice capability and exhibited subpar batch correction. Mask-GAN demonstrated some improvement in batch correction but with limited efficacy. On the other hand, Mask-G2N notably boosted model performance, although it fell short of STG3Net. This outcome suggests promising prospects for enhancing results through combining adversarial learning and G2N anchor pairs selection.

Next, we examined the impacts of various objective functions in STG3Net. We explored scaled cosine loss (SCE) and mean squared error (MSE) for masked gene reconstruction. Additionally, for selected G2N-pairs, we evaluated the effects of triplet loss and contrastive loss to optimize anchor distances. The results demonstrated that STG3Net achieved the highest accuracy and batch correction performance by employing SCE and triplet loss.

Finally, we explored the influence of mask rates and the number of anchor pairs on overall performance. For each set of values, we ran 10 trials with different seeds, calculated the median values for the 12 slices, and computed the mean and standard deviation (\autoref{fig-4}B).

\section{CONCLUSIONS AND DISCUSSION}\label{CONCLUSIONS}
In this paper, we propose STG3Net, a novel method designed to integrate multiple SRT datasets and correct batch effects. We enhance the robustness of the self-supervised encoder by reconstructing masked spot expressions, which helps in reducing the occurrence of discrete spots. We employ adversarial learning to improve the model's ability to identify spatial domains in complex, multi-slice SRT data. Additionally, we introduce a plug-and-play batch correction method called Global Nearest Neighbor (G2N) anchor pair selection, specifically designed for SRT datasets. G2N facilitates faster acquisition of anchor pairs by considering global semantic information and minimizing the selection of erroneous pairs. We validate STG3Net on three diverse datasets using multiple evaluation metrics, including continuity, accuracy, and effectiveness of batch correction. For the DLPFC donor3 data, the latent representation generated by STG3Net enables UMAP visualization, effectively capturing cortical layers and blending multiple slices. The performance on AMB and ME datasets demonstrates STG3Net's ability to preserve variability and connectivity during biological development. Furthermore, an ablation study is conducted to explore the model's performance by examining the impact of different components, objective functions, and other factors from various perspectives.


\section*{Acknowledgment}
The work was supported in part by the National Natural Science Foundation of China (62262069), in part by the Yunnan Fundamental Research Projects under Grant (202201AT070469) and the Yunnan Talent Development Program - Youth Talent Project. 

\balance
\small
\bibliography{references.bib} 
\bibliographystyle{IEEEtran}  
\end{document}